%% file: paper.tex
\documentclass[sigconf]{acmart}
\renewcommand\footnotetextcopyrightpermission[1]{} 

\usepackage{booktabs}

\usepackage{url,bm}
\usepackage{latexsym}

\usepackage{anyfontsize}

\usepackage{algorithm}
\usepackage[noend]{algorithmic}
\usepackage{multirow}
\usepackage{amssymb}

\usepackage{etoolbox}
\newtoggle{conf}
\toggletrue{conf}

\newcommand{\cut}[1]{}

\begin{document}

\title{Towards Deep and Representation Learning\\for Talent Search at LinkedIn}

\author{Rohan Ramanath $^*$, Hakan Inan $^{*,+}$, Gungor Polatkan $^*$, Bo Hu, Qi Guo, Cagri Ozcaglar\\Xianren Wu, Krishnaram Kenthapadi, Sahin Cem Geyik $^*$}

\affiliation{
  \institution{LinkedIn Corporation, USA}
}

\renewcommand{\shortauthors}{Ramanath et al.}

\begin{abstract}
\input{abstract}
\end{abstract}


\setcopyright{none}

\maketitle

\input{macros}

\input{intro}
\input{problem}
\input{methodology}

\vspace{-0.05in}
\input{exp}

\input{lessons}
\vspace{-0.07in}
\input{related}

\input{conclusion}

{
	\scriptsize
	\bibliographystyle{abbrv}
	\bibliography{paper}
}

\end{document}

%% file: abstract.tex
Talent search and recommendation systems at LinkedIn strive to match the potential candidates to the hiring needs of a recruiter or a hiring manager expressed in terms of a search query or a job posting. Recent work in this domain has mainly focused on linear models, which do not take complex relationships between features into account, as well as ensemble tree models, which introduce non-linearity but are still insufficient for exploring all the potential feature interactions, and strictly separate feature generation from modeling. In this paper, we present the results of our application of deep and representation learning models on LinkedIn Recruiter. Our key contributions include:  (\emph{i}) Learning semantic representations of sparse entities within the talent search domain, such as recruiter ids, candidate ids, and skill entity ids, for which we utilize neural network models that take advantage of LinkedIn Economic Graph, and (\emph{ii}) Deep models for learning recruiter engagement and candidate response in talent search applications. We also explore learning to rank approaches applied to deep models, and show the benefits for the talent search use case. Finally, we present offline and online evaluation results for LinkedIn talent search and recommendation systems, and discuss potential challenges along the path to a fully deep model architecture. The challenges and approaches discussed generalize to any multi-faceted search engine.

%% file: macros.tex
%
%
\def\x{\mathbf{x}}
\def\w{\mathbf{w}}
\def\sp{s^{(p)}}
\def\pp{{i^+}}
\def\nn{{i^-}}
\def\minimize{\text{minimize}}
\def\hess{{\boldsymbol \nabla}^2}
\newcommand{\eq}[1]{\begin{alignat}{3}#1\end{alignat}}
\newcommand{\eqn}[1]{\begin{alignat*}{3}#1\end{alignat*}}
\newcommand{\commentout}[1]{}
\newcommand{\dotprod}[2]{\left\langle#1,#2\right\rangle}
\newcommand{\norm}[1]{\left\|#1\right\|_2}
\newcommand{\inner}[1]{{\langle #1 \rangle}}
\newcommand{\widesim}[2][1.5]{
  \mathrel{\overset{#2}{\scalebox{#1}[1]{$\sim$}}}
}

\newcommand{\murat}[1]{{\color{magenta}[MURAT: #1]}}

%% file: intro.tex
\section{Introduction}\label{sec:intro}

LinkedIn Talent Solutions business contributes to around 65\% of LinkedIn's annual revenue$^1$\let\thefootnote\relax\footnote{$^1$ ~ https://press.linkedin.com/about-linkedin}, and provides tools for job providers to reach out to potential candidates and for job seekers to find suitable career opportunities. LinkedIn's job ecosystem has been designed as a platform to connect job providers and job seekers, and to serve as a marketplace for efficient matching between potential candidates and job openings. A key mechanism to help achieve these goals is the \emph{LinkedIn Recruiter} product, which enables recruiters to search for relevant candidates and obtain candidate recommendations for their job postings.

\footnote{$^*$ ~ Authors contributed equally to this work.} \footnote{$^+$ ~ Current affiliation: Stanford University. Work done during an internship at LinkedIn.} \footnote{\\ ~ \\ ~  \large \textbf{This paper has been accepted for publication at ACM CIKM 2018.}}A crucial challenge in talent search and recommendation systems is that the underlying query could be quite complex, combining several structured fields (such as canonical title(s), canonical skill(s), company name) and unstructured fields (such as free-text keywords). Depending on the application, the query could either consist of an explicitly entered query text and selected facets (talent search), or be implicit in the form of a job opening, or ideal candidate(s) for a job (talent recommendations). Our goal is to determine a ranked list of most relevant candidates among hundreds of millions of structured candidate profiles.

The structured fields add sparsity to the feature space when used as a part of a machine learning ranking model. This setup lends itself well to a dense representation learning experiment as it not only reduces sparsity but also increases sharing of information in feature space. In this work, we present the experiences of applying representation learning techniques for talent search ranking at LinkedIn. Our key contributions include:
\vspace{-\topsep}
\begin{itemize}
\item Using embeddings as features in a learning to rank application. This typically consists of:
\begin{itemize}
\item Embedding models for ranking, and evaluating the advantage of a layered (fully-connected) architecture,
\item Considerations while using point-wise learning and pair-wise losses in the cost function to train models.
\end{itemize}
\item Methods for learning semantic representations of sparse entities (such as recruiter id, candidate id, and skill id) using the structure of the LinkedIn Economic Graph~\cite{Wei12}:
\begin{itemize}
\item Unsupervised representation learning that uses Economic Graph network data across LinkedIn ecosystem
\item Supervised representation learning that utilizes application specific data from talent search domain.
\end{itemize}
\item Extensive offline and online evaluation of above approaches in the context of LinkedIn talent search, and a discussion of challenges and lessons learned in practice.
\end{itemize}
\vspace{-\topsep}
Rest of the paper is organized as follows. We present an overview of talent search at LinkedIn, the constraints, challenges and the optimization problem in \S\ref{sec:problem}. We then describe the methodology for the application of representation learning models in \S\ref{sec:methodology}, followed by offline and online experimentation results in \S\ref{sec:exp}. Finally, we discuss related work in \S\ref{sec:related}, and conclude the paper in \S\ref{sec:conclusion}.

Although much of the discussion is in the context of search at LinkedIn, it generalizes well to any multi-faceted search engine where there are high dimensional facets, i.e. movie, food / restaurant, product search are a few examples that would help the reader connect to the scale of the problem.

%% file: problem.tex
\section{Background and Problem Setting}\label{sec:problem}
We next provide a brief overview of the {\em LinkedIn Recruiter} product and existing ranking models, and formally present the talent search ranking problem.

\subsection{Background}
\begin{figure}[t]
\centering
\includegraphics[width=3.3in]{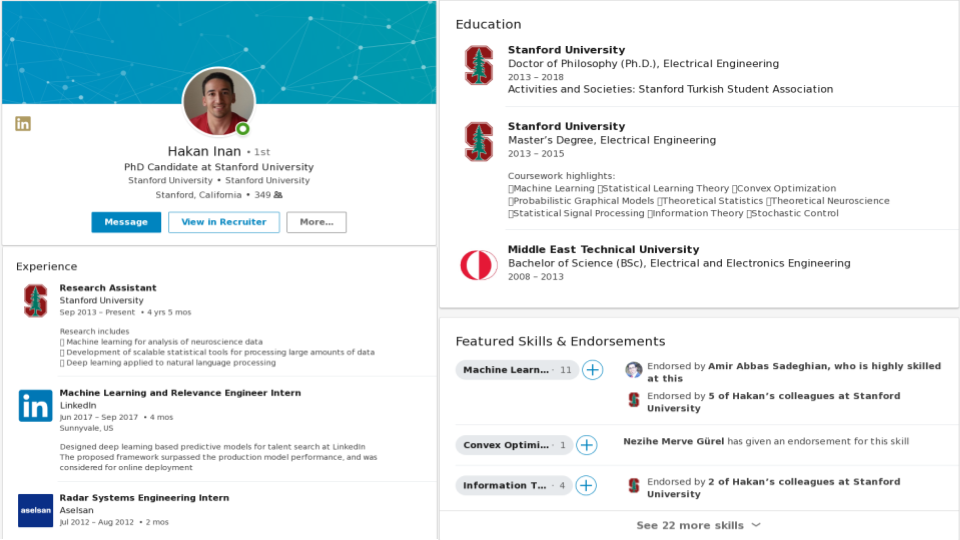}
\vspace{-0.05in}
s\caption{An example LinkedIn profile page}
\label{fig:profile}
\vspace{-0.1in}
\end{figure}

LinkedIn is the world's largest professional network with over 500 million members world-wide. Each member of LinkedIn has a profile page that serves as a professional record of achievements and qualifications, as shown in Figure~\ref{fig:profile}. A typical member profile contains around 5 to 40 structured and unstructured fields including title, company, experience, skills, education, and summary, amongst others.

In the context of talent search, LinkedIn members can be divided into two categories: candidates (job seekers) and recruiters (job providers). Candidates look for suitable job opportunities, while recruiters seek candidates to fill job openings. In this work, we address the modeling challenges in the \emph{LinkedIn Recruiter} product, which helps recruiters find and connect with the right candidates.
 
\begin{figure}[t]
\centering
\includegraphics[width=3.3in]{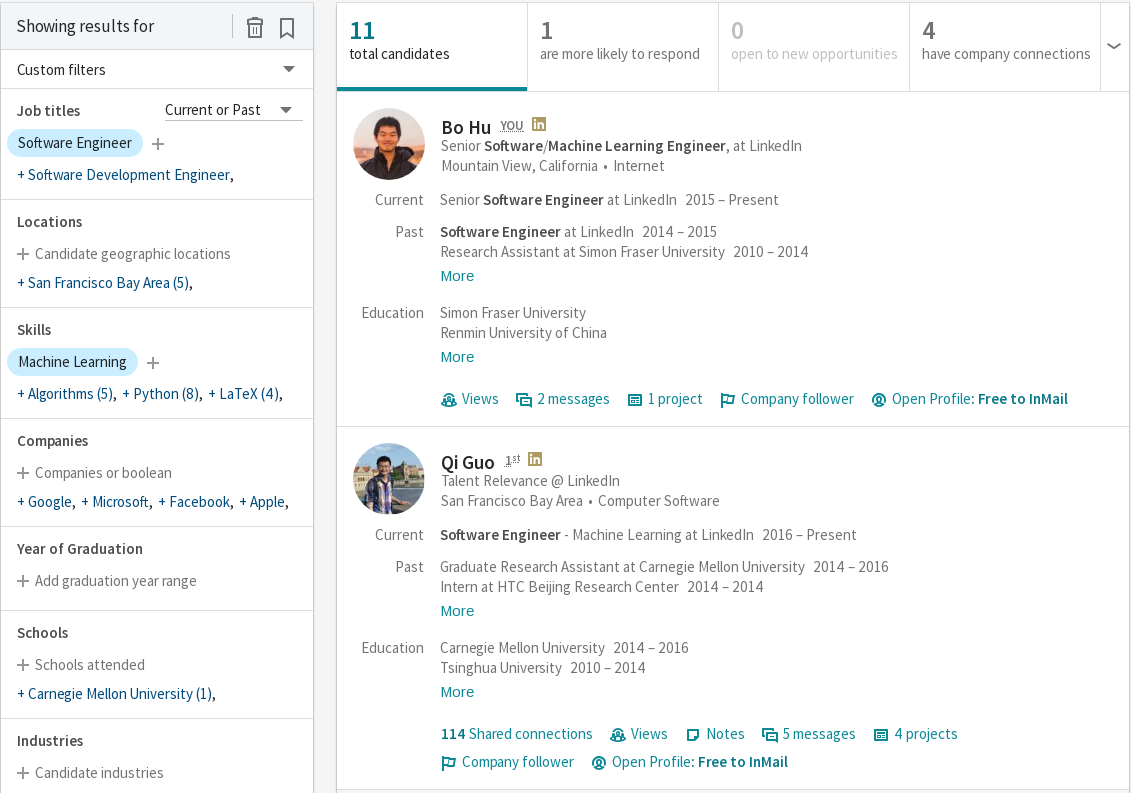}
\vspace{-0.05in}
\caption{A recruiter seeks a software engineer with background in machine learning}
\label{fig:serp}
\vspace{-0.1in}
\end{figure}

Consider an example of a recruiter looking for a software engineer with machine learning background. Once the recruiter types keywords \emph{software engineer} and \emph{machine learning} as a free text query, the recruiter search engine first standardizes them into the title, \emph{software engineer} and the skill, \emph{machine learning}. Then, it matches these standardized entities with standardized member profiles, and the most relevant candidate results are presented as in Figure~\ref{fig:serp}. In order to find a desired candidate, the recruiter can further refine their search criteria using facets such as title, location, and industry. For each result, the recruiter can perform the following actions (shown in the increasing order of recruiter's interest for the candidate):
\begin{enumerate}
\item View a candidate profile,
\item Bookmark a profile for detailed evaluation later,
\item Save a profile to their current hiring project (as a potential fit), and,
\item Send an inMail (message) to the candidate.
\end{enumerate}
Unlike traditional search and recommendation systems which solely focus on estimating how relevant an item is for a given query, the talent search domain requires mutual interest between the recruiter and the candidate in the context of the job opportunity. In other words, we simultaneously require that a candidate shown must be relevant to the recruiter's query, and that the candidate contacted by the recruiter must also show interest in the job opportunity. Therefore, we define a new action event, \emph{inMail Accept}, which occurs when a candidate replies to an inMail from a recruiter with a positive response. Indeed, the key business metric in the Recruiter product is based on inMail Accepts and hence we use the fraction of top $k$ ranked candidates that received and accepted an inMail (viewed as \emph{precision@k}$~^2$\footnote{$^2$ ~ While metrics like \emph{Normalized Discounted Cumulative Gain} \cite{jarvelin_2002} are more commonly utilized for ranking applications, we have found precision@k to be more suitable as a business metric. Precision@k also aligns well with the way the results are presented in LinkedIn Recruiter application, where each page lists up to 25 candidates by default so that precision@25 is a measure of positive outcome in the first page.}) as the main evaluation measure for our experiments.

\subsection{Current Models} \label{subsec:currentmodels}
The current talent search ranking system functions as follows~\cite{Thuc16,thuc15pes}. In the first step, the system retrieves a candidate set of a few thousand members from over 500 million LinkedIn members, utilizing hard filters specified in the search query. In particular, a query request is created based on the standardized fields extracted from the free form text in the query, as well as the selected facets (such as skill, title, and industry). This query request is then issued to the distributed search service tier, which is built on top of LinkedIn's \emph{Galene} search platform~\cite{galene_engine}. A list of candidates is generated based on the matching features (such as title or skill match). In the second step, the search ranker scores the resulting candidates using a ranking model, and returns the top ranked candidate results. In this paper, we focus on the ranking model used in the second step.

Tree ensembles, viewed as non-linear models capable of discovering ``interaction features'', have been studied extensively both in academic and industrial settings. For LinkedIn Talent Search, we have studied the effectiveness of ensemble trees and found that GBDT (Gradient Boosted Decision Trees) models \cite{friedman_2001,xgboost} outperform the best logistic regression models on different data sets as measured by area under ROC curve (AUC) and precision@k metrics in offline experiments. More importantly, online A/B tests have shown significant improvement across all key metrics based on inMail Accepts (Table~\ref{tab:linearvsxgboost}). Having described the current state of the talent search ranking models, we next formally present the problem setting and the associated challenges.

\begin{table}[!ht]
\small
\caption{Results of an online A/B test we performed over a period of three weeks in 2017, demonstrating the precision improvement for the gradient boosted decision tree model compared to the baseline logistic regression model for LinkedIn Talent Search. We compute precision as the fraction of the top ranked candidate results that received and accepted an inMail, within three days of the inMail being sent by the recruiter (\emph{Prec@k} is over the top $k$ candidate results), and show the relative lift in precision. We note that these improvements are impressive based on our experience in the domain.}
\vspace{-0.15in}
\begin{tabular}{c|c|c|c}
\hline\hline
\textbf{ } & \textbf{Prec@5} & \textbf{Prec@25} & \textbf{Overall precision} \\\hline\hline
Lift & +7.5\% & +7.4\%  & +5.1\% \\\hline
p-value & 2.1e-4 & 4.8e-4  & 0.01 \\\hline
\end{tabular}
\label{tab:linearvsxgboost}
\vspace{-0.1in}
\end{table}

\subsection{Problem Setting and Challenges}
\begin{definition}
Given a search query consisting of search criteria such as title, skills, and location, provided by the recruiter or the hiring manager, the goal of \emph{Talent Search Ranking} is to:
\begin{enumerate}
\item Determine a set of candidates strictly satisfying the specified search criteria (hard constraints), and,
\item Rank the candidates according to their utility for the recruiter, where the utility is the likelihood that the candidate would be a good fit for the position, and would be willing to accept the request (inMail) from the recruiter.
\end{enumerate}
\end{definition}
As discussed in \S\ref{subsec:currentmodels}, the existing ranking system powering \\LinkedIn Recruiter product utilizes a GBDT model due to its advantages over linear models. While GBDT provides quite a strong performance, it poses the following challenges:
\begin{enumerate}
\item It is quite non-trivial to augment a tree ensemble model with other trainable components such as embeddings for discrete features. Such practices typically require joint training of the model with the component/feature, while the tree ensemble model assumes that the features themselves need not be trained.
\item Tree models do not work well with sparse id features such as skill ids, company ids, and member ids that we may want to utilize for talent search ranking. Since a sparse feature is non-zero for a relatively small number of examples, it has a small likelihood to be chosen by the tree generation at each boosting step, especially since the learned trees are shallow in general.
\item Tree models lack flexibility of model engineering. It might be desirable to use novel loss functions, or augment the current objective function with other terms. Such modifications are not easily achievable with GBDT models, but are relatively straight-forward for deep learning models based on differentiable programming. A neural network model with a final (generalized) linear layer also makes it easier to adopt approaches such as transfer learning and online learning.
\end{enumerate}
In order to overcome these challenges, we explore the usage of neural network based models, which provide sufficient flexibility in the design and model specification.

Another significant challenge pertains to the sheer number of available entities that a recruiter can include as part of their search, and how to utilize them for candidate selection as well as ranking. For example, the recruiter can choose from tens of thousands of LinkedIn's standardized skills. Since different entities could be related to each other (to a varying degree), using syntactic features (e.g., fraction of query skills possessed by a candidate) has its limitations. Instead, it is more desirable to utilize semantic representations of entities, for example, in the form of low dimensional embeddings. Such representations allow for numerous sparse entities to be better incorporated as part of a machine learning model. Therefore, in this work, we also investigate the application of representation learning for entities in the talent search domain.

%% file: methodology.tex
\section{Methodology} \label{sec:methodology}
In this section, we present our methodology which focuses on two main aspects:
\begin{itemize}
\item Learning of deep models to estimate likelihood of the two-way interest (inMail accept) between the candidate and the recruiter,
\item Learning of supervised and unsupervised embeddings of the entities in the talent search domain.
\end{itemize}
In Table~\ref{tab:notations}, we present the notation used in the rest of the section. Note that the term $example$ refers to a candidate that is presented to the recruiter.
\begin{table}[!ht]
\small
\caption{Notations}
\vspace{-0.15in}
\begin{tabular}{c|c}
\hline\hline
\textbf{Notation} & \textbf{Represents} \\ \hline\hline
$n$ & size of a training set \\ \hline
$\x_i$ &feature vector for $i^\text{th}$ example  \\ \hline
$y_i$ & binary response for $i^\text{th}$ example \\ & (inMail accept or not)  \\ \hline
$s_i$ & the search session to which \\ & $i^\text{th}$ example belongs \\ \hline
tuple $(\x_i, y_i, s_i)$ & $i^\text{th}$ example in the training set \\ \hline
$\w$ & weight vector \\ \hline
$\dotprod{\cdot}{\cdot}$ & dot product \\ \hline
$\psi(\cdot)$ & neural network function \\ \hline
$u_j \in \mathbb{R}^d$ & $d$ dimensional \\ & vector representation of entity j
\\ \hline \hline
\end{tabular}
\label{tab:notations}
\vspace{-0.12in}
\end{table}

\input{methodology_subsec_1}

\input{methodology_subsec_2}

\input{methodology_subsec_3}

%% file: methodology_subsec_1.tex
\subsection{Embedding Models for Ranking} \label{subsec:deepmodel} 

As mentioned before, we would like to have a flexible ranking model that allows for easy adaptation to novel features and training schemes. Neural networks, especially in the light of recent advances that have made them the state of the art for many statistical learning tasks including learning to rank \cite{burges_2005,tyliu_2009}, are the ideal choice owing to their modular structure and their ability to be trained end-to-end using straightforward gradient based optimization. Hence we would like to use neural network rankers as part of our ranking models for Talent Search at LinkedIn. Specifically, we propose to utilize multilayer perceptron (MLP) with custom unit activations for the ranking task. Our model supports a mix of model regularization methods including L2 norm penalty and dropout \cite{srivastava_2014}.

For the  training objective of the neural network, we consider two prevalent methodologies used in learning to rank:
\vspace{-2mm}
\subsubsection{Pointwise Learning} Also called \emph{ranking by binary classification}, this method involves training a binary classifier utilizing each example in the training set with their labels, and then grouping the examples from the same search session together and ranking them based on their scores. For this purpose, we apply logistic regression on top of the neural network as follows. We include a classification layer which sums the output activations from the neural network, passes the sum through the logistic function, and then trains against the labels using the cross entropy cost function:
\begin{equation}
\sigma_i = \frac{1}{1+\exp\left(-\dotprod{\w}{\psi(\x_i)}\right)}, \hspace{5mm} i \in \{1, \cdots, n\}
\end{equation}
\begin{equation}
\mathcal{L} = -\sum_{i=1}^n y_i \log(\sigma_i) + (1-y_i) \log(1 - \sigma_i)
\end{equation}
In above equations, $\psi(\cdot)$ refers to the neural network function, and $\sigma_i$ is the value of the logistic function applied to the score for the $i^\text{th}$ training example.
\vspace{-2mm}
\subsubsection{Pairwise Learning} Although pointwise learning is simple to implement and works reasonably well, the main goal for talent search ranking is to provide a ranking of candidates which is guided by the information inherent in available session-based data. Since it is desirable to compare candidates within the same session depending on how they differ with respect to the mutual interest between the recruiter and the candidate (inMail accept), we form pairs of examples with positive and negative labels respectively from the same session and train the network to maximize the difference of scores between the paired positive and negative examples:
\begin{align}
d_{\pp,\nn} &= \dotprod{\w}{\psi(\x_\pp)-\psi(\x_\nn)}, \\
\label{eq:pairwise_loss}
\mathcal{L} &=  \sum_{\footnotesize \begin{array}{c} (\pp, \nn) : s_{\pp} = s_{\nn}, \\ y_{\pp}=1,  y_{\nn}=0\end{array}} f(d_{\pp,\nn}).
\end{align}
The score difference between a positive and a negative example is denoted by $d_{\pp,\nn}$, with $\pp$ and $\nn$ indicating the indices for a positive and a negative example, respectively. The function $f(\cdot)$ determines the loss, and \eqref{eq:pairwise_loss} becomes equivalent to the objective of RankNet \cite{burges_2005} when $f$ is the logistic loss:
\begin{align*}
f(d_{\pp,\nn}) = \log\left(1 + \exp(-d_{\pp,\nn})\right),
\end{align*}
whereas \eqref{eq:pairwise_loss} becomes equivalent to ranking SVM objective \cite{joachims_2002} when $f$ is the hinge loss:
\begin{align*}
f(d_{\pp,\nn}) = \max(0, 1-d_{\pp,\nn}).
\end{align*}
We implemented both pointwise and pairwise learning objectives. For the latter, we  chose hinge loss over logistic loss due to faster training times, and our observation that the precision values did not differ significantly (we present the evaluation results for point-wise and hinge loss based pairwise learning in \S\ref{sec:exp}).

%% file: methodology_subsec_2.tex
\subsection{Learning Semantic Representations of Sparse Entities in Talent Search} \label{subsec:representation}
Next, we would like to focus on the problem of sparse entity representation, which allows for the translation of the various entities (skills, titles, etc.) into a low-dimensional vector form. Such a translation makes it possible for various types of models to directly utilize the entities as part of the feature vector, e.g., \S\ref{subsec:deepmodel}. To achieve this task of generating vector representations, we re-formulate the talent search problem as follows: given a query \(q\) by a recruiter \(r_i\), rank a list of LinkedIn members  \(m_1,m_2,...,m_d\) in the order of decreasing relevance. In other words, we want to learn a function that assigns a score for each potential candidate, corresponding to the query issued by the recruiter. Such a function can learn a representation for query and member pair, and perform final scoring afterwards. We consider two broad approaches for learning these representations.
\begin{itemize}
\item The \textbf{\emph{unsupervised approach}} learns a shared representation space for the entities, thereby constructing a query representation and a member representation. We do not use talent search specific interactions to supervise the learning of representations.
\item The \textbf{\emph{supervised approach}} utilizes the interactions between recruiters and candidates in historical search results while learning both representation space as well as the final scoring.
\end{itemize}

\begin{figure}
	\includegraphics[width=2.7in]{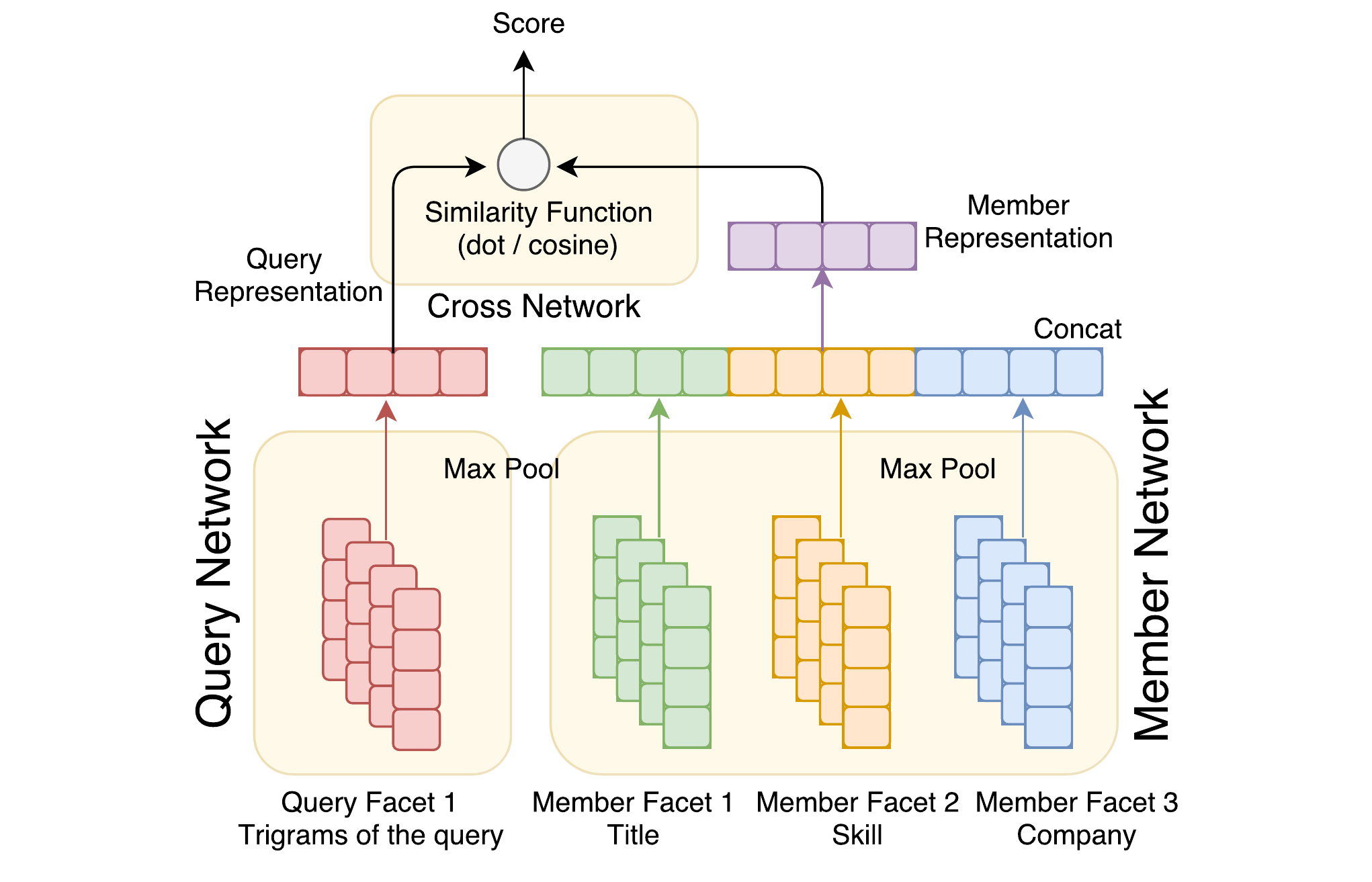}
	\vspace{-0.1in}
	\caption{The two arm architecture with a shallow query arm and a deep member arm}
	\label{fig:model}
	\vspace{-0.1in}
\end{figure}

The architecture for learning these representations and the associated models is guided by the need to scale for deployment to production systems that serve over 500M members. For this reason, we split the network scoring of query-member pair into three semantic pieces, namely query network, member network, and cross network, such that each piece is run on one of the production systems as given in Figure \ref{fig:model}.

\subsubsection{Unsupervised Embeddings} \label{subsubsec:unsupervised_emb}
~~~~ Most features used in \\LinkedIn talent search and recommendation models are categorical in nature, representing entities such as skill, title, school, company, and other attributes of a member's profile. In fact, to achieve personalization, even the member herself could be represented as a categorical feature via her LinkedIn member Id. Such categorical features often suffer from sparsity issues because of the large search space, and learning a dense representation to represent these entities has the potential to improve model performance. While commonly used algorithms such as \textit{word2vec}~\cite{Mikolov13} work well on text data when there is information encoded in the sequence of entities, they cannot be directly applied to our use case. Instead, we make use of LinkedIn Economic Graph~\cite{Wei12} to learn the dense entity representations.

LinkedIn Economic Graph is a digital representation of the global economy based on data generated from over $500$ million members, tens of thousands of standardized skills, millions of employers and open jobs, as well as tens of thousands of educational institutions, along with the relationships between these entities. It is a compact representation of all the data on LinkedIn. To obtain a representation for the entities using the Economic Graph, we could use a variety of graph embedding algorithms (see \S\ref{sec:related}). For the purposes of this work, we adopted \emph{Large-Scale Information Network Embeddings} approach~\cite{Tang15}, by changing how we construct the graph. In~\cite{Tang15}, the authors construct the graph of a social network by defining the members of the network as vertices, and use some form of interaction (clicks, connections, or social actions) between members to compute the weight of the edge between any two members. In our case, this would create a large sparse graph resulting in intractable training and a noisy model. Instead, we define a weighted graph, $G = (V, E, w_{..})$ over the entities whose representations need to be learned (e.g., skill, title, company), and use the number of members sharing the same entity on their profile to induce an edge weight ($w_{..}$) between the vertices. Thus we reduce the size of the problem by a few orders of magnitude by constructing a smaller and denser graph.

\begin{figure}
	\includegraphics[width=2in]{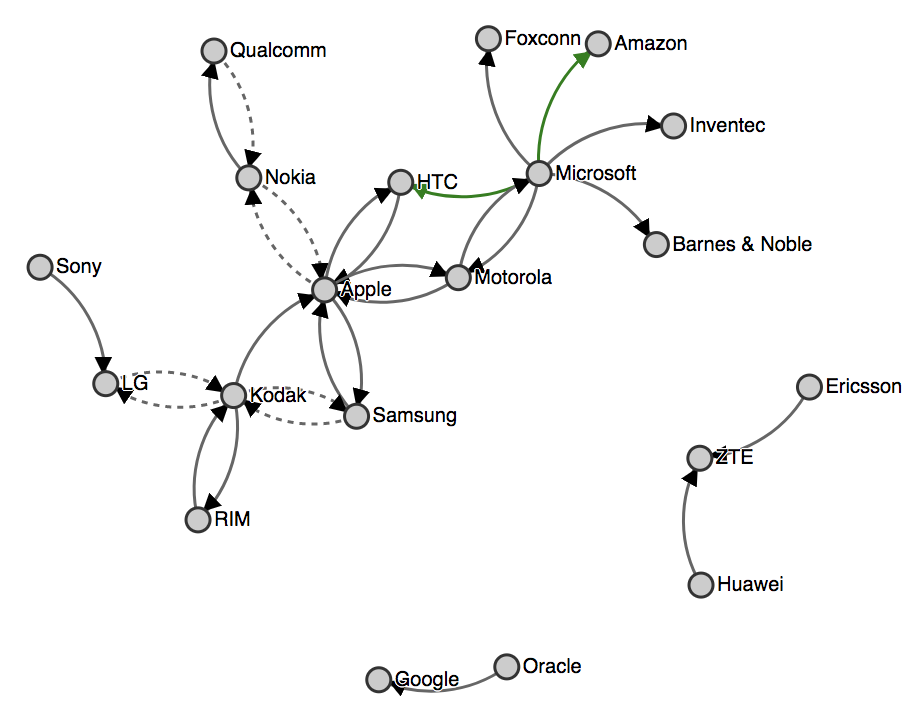}
	\vspace{-0.1in}
	\caption{Each vertex represents a company; the edge weight denoted by color, dashed or regular edge represents \#members that have worked at both companies.} 
	\label{fig:CompanyNetworkEmbeddings}
	\vspace{-0.1in}
\end{figure}

An illustrative sub-network of the graph used to construct company embeddings is presented in Figure \ref{fig:CompanyNetworkEmbeddings}. Each vertex in the graph represents a company, and the edge weight (denoted by the edge thickness) represents the number of LinkedIn members that have worked at both companies (similar graphs can be constructed for other entity types such as skills and schools). In the example, our aim would be to embed each company (i.e., each vertex in the graph) into a fixed dimensional latent space. We propose to learn \emph{first order} and \emph{second order} embeddings from this graph. 
Our approach, presented below, is similar to the one proposed in~\cite{Tang15}.

\textbf{First order embeddings}
Corresponding to each undirected edge between vertices $v_i$ and $v_j$, we define the joint probability between vertices $v_i$ and $v_j$ as:
\begin{equation}
p_1(v_i,v_j) = \frac{1}{Z} \cdot \frac{1}{ 1+ exp(- \dotprod{u_i}{u_j})} ~ ,
\end{equation}
where $u_i \in \mathbb{R}^d$ is the d-dimensional vector representation of vertex $v_i$ and $Z = \sum_{(v_i, v_j) \in E} \frac{1}{ 1+ exp(- \dotprod{u_i}{u_j})}$ is the normalization factor. The empirical probability, $\hat{p}_1(\cdot,\cdot)$ over the space $V \times V$ can be calculated using:
\begin{equation}
\label{eq:empirical_p1}
\hat{p}_1(v_i, v_j) = \frac{w_{ij}}{W} ~ ,
\end{equation}
where $w_{ij}$ is the edge weight in the company graph, and $W = \displaystyle\sum_{ (v_i, v_j) \in E } w_{ij}$. We minimize the following objective function in order to preserve first-order proximity:
\begin{equation}
O_1 = d( \hat{p}_1(\cdot,\cdot), p_1(\cdot,\cdot) ) ~ ,
\end{equation}
where $d(\cdot,\cdot)$ is a measure of dissimilarity between two probability distributions. We chose to minimize KL-divergence of $\hat{p}_1$ with respect to $p_1$:
\begin{equation}
O_1 = - \sum_{ (v_i, v_j) \in E } \hat{p}_1(v_i, v_j) \log \bigg( \frac{p_1(v_i, v_j)}{\hat{p}_1(v_i, v_j)} \bigg) ~ .
\end{equation}

\textbf{Second order embeddings}
Second order embeddings are generated based on the observation that vertices with shared neighbors are similar. In this case, each vertex plays two roles: the vertex itself, and a specific context of other vertices. Let $u_i$ and ${u_i}^{\prime}$ be two vectors, where $u_i$ is the representation of $v_i$ when it is treated as a vertex, while ${u_i}^{\prime}$ is the representation of $v_i$ when it is used as a specific context. For each directed edge $(i,j)$, we define the probability of context $v_j$ to be generated by vertex $v_i$ as follows:
\begin{equation}
p_2(v_j | v_i) = \frac{exp(\dotprod{u'_j}{u_i})}{\displaystyle\sum_{k=1}^{|V|} exp(\dotprod{u'_k}{u_i})} ~ .
\end{equation}
The corresponding empirical probability can be obtained as:
\begin{equation}
\label{eq:empirical_p2}
\hat{p}_2(v_j | v_i) = \frac{w_{ij}}{W_i} ~ ,
\end{equation}
where $W_i = \displaystyle\sum_{ v_j: (v_i, v_j) \in E } w_{ij}$.
In order to preserve the second order proximity, we aim to make conditional probability distribution of contexts, $p_2(\cdot|v_i)$, to be close to empirical probability distribution $\hat{p}_2(\cdot|v_i)$, by minimizing the following objective function:
\begin{equation}
O_2 = \sum_{v_i \in V} \lambda_i \cdot d (\hat{p}_2(\cdot | v_i), p_2(\cdot | v_i)) ~ ,
\end{equation}
where $d(\cdot, \cdot)$ is a measure of dissimilarity between two probability distributions, and $\lambda_i$ represents the importance of vertex $v_i$ (e.g., computed using PageRank algorithm). In this work, for simplicity, we set $\lambda_i$ to be the degree of vertex $v_i$. Using KL-divergence as before, the objective function for the second order embeddings can be rewritten as:
\begin{equation}
O_2 = \sum_{v_i \in V} \lambda_i \cdot 
\sum_{ v_j: (v_i, v_j) \in E } \hat{p}_2(v_j | v_i) \log \bigg( \frac{p_2(v_j|v_i)}{\hat{p}_2(v_j | v_i)} \bigg) ~ .
\end{equation}

Using Figure \ref{fig:CompanyNetworkEmbeddings}, we can now explain how the feature is constructed for each member. After optimizing for $O_1$ and $O_2$ individually using gradient descent, we now have two vectors for each vertex of the graph (i.e. in this case the company). A company can now be represented as a single vector by concatenating the first and second order embeddings. This represents each company on a single vector space. Each query and member can be represented by a bag of companies, i.e. a query can contain multiple companies referenced in the search terms and a member could have worked at multiple companies which is manifested on the profile. Thus, with a simple pooling operation (max-pooling or mean-pooling) over the bag of companies, we can represent each query and member as a point on the vector space. A similarity function between the two vector representations can be used as a feature in ranking.

\subsubsection{Supervised Embeddings} \label{subsubsec:supervised_emb}
In this section, we explain how to train the entity embeddings in a supervised manner. We first collect the training data from candidates recommended to the recruiters (with the inMail accept events as the positive labels) within the LinkedIn Recruiter product, and then learn the feature representations for the entities guided by the labeled data. For this purpose, we adopted and extended \emph{Deep Semantic Structured Models} (DSSM) based learning architecture~\cite{Huang13}. In this scheme, document and query text are modeled through separate neural layers and crossed before final scoring, optimizing for the search engagement as a positive label. Regarding features, the DSSM model uses the query and document text and converts them to character trigrams, then utilizes these as inputs to the model. An example character trigram of the word \emph{java} is given as  \{\#ja, jav, ava, va\#\}. This transformation is also called \emph{word-hashing} and instead of learning  a vector representation (i.e. embedding) for the entire word, this technique provides representations for each character trigram. In this section we extend this scheme and add categorical representations of each type of entity as inputs to the DSSM model.

We illustrate our usage of word-hashing through an example. Suppose that a query has the title id $t_i$ selected as a facet, and contains the search box keyword, \emph{java}. We process the text to generate the following trigrams:  \{\#ja, jav, ava, va\#\}. Next, we add the static standardized ids corresponding to the selected entities ($t_i$, in this example) as inputs to the model. We add entities from the facets to the existing model, since text alone is not powerful enough to encode the semantics. After word hashing, a multi-layer non-linear projection (consisting of multiple fully connected layers) is performed to map the query and the documents to a common semantic representation. Finally, the similarity of the document to the query is calculated using a vector similarity measure (e.g., cosine similarity) between their vectors in the new learned semantic space. We use stochastic gradient descent with back propagation to infer the network coefficients. 

Output of the model is a set of representations (i.e. dictionary) for each entity type (e.g. title, skill) and network architecture that we re-use during the inference. Each query and member can be represented by a bag of entities, i.e. a query can contain multiple titles and skills referenced in the search terms and a member could have multiple titles and skills which are manifested on the profile. The lookup tables learned during training  and network coefficients are used to construct query and document embeddings. The two arms of DSSM corresponds to the supervised embeddings of the query and the document respectively. We then use the similarity measured by the distance of these two vectors (e.g. cosine) as a feature in the learning to rank model.

We used DSSM models over other deep learning models and nonlinear models for the following reasons. First, DSSM enables projection of recruiter queries (query) and member profiles (document) into a common low-dimensional space, where relevance of the query and the document can be computed as the distance between them. This is important for talent search models, as the main goal is to find the match between recruiter queries and member profiles. Secondly, DSSM uses word hashing, which enables handling large vocabularies, and results in a scalable semantic model.

%% file: methodology_subsec_3.tex
\subsection{Online System Architecture} 
\label{subsec:arch}
\begin{figure}[t]
\centering
\includegraphics[width=3.0in]{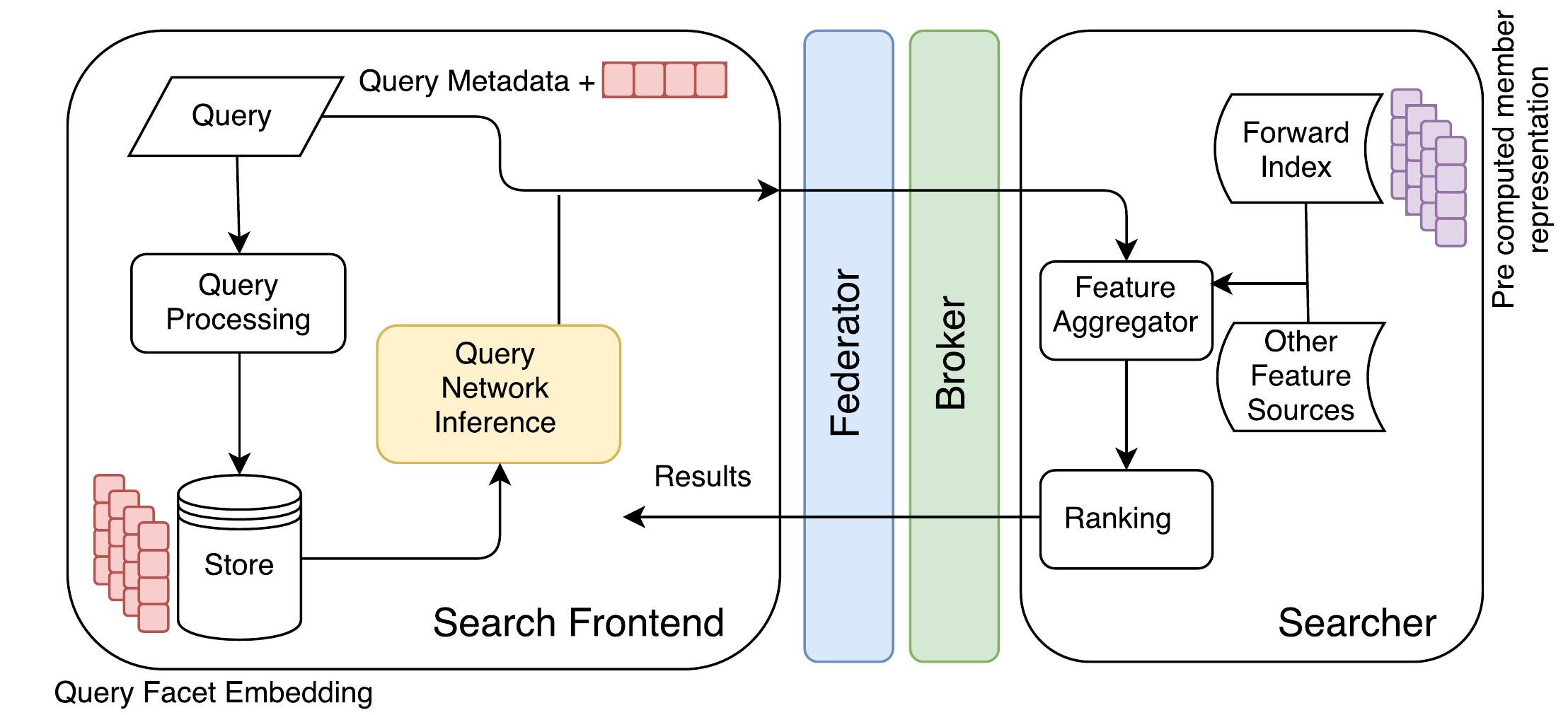}
\vspace{-0.1in}
\caption{Online System Architecture for Search Ranking}
\label{fig:arch}
\end{figure}

Figure~\ref{fig:arch} presents the online architecture of the proposed talent search ranking system, which also includes the embedding step (\S\ref{subsec:representation}). We designed our architecture such that the member embeddings are computed offline, but the query embeddings are computed at run time. We made these choices for the following reasons: (1) since a large number of members may match a query, computing the embeddings for these members at run time would be computationally expensive, and, (2) the queries are typically not known ahead of time, and hence the embeddings need to be generated online. Consequently, we chose to include member embeddings as part of the forward index containing member features, which is generated periodically by an offline workflow (not shown in the figure). We incorporated the component for generating query embeddings as part of the online system. Our online recommendation system consists of two services:
\begin{enumerate}
\item \textbf{Retrieval Service:} This service receives a user query, generates the candidate set of members that match the criteria specified in the query, and computes an initial scoring of the retrieved candidates using a simple, first-pass model. These candidates, along with their features, are retrieved from a distributed index and returned to the scoring/ranking service. The features associated with each member can be grouped into two categories:
\begin{itemize}
\item \emph{Explicit Features:} These features correspond to fields that are present in a member profile, e.g., current and past work positions, education, skills, etc.
\item \emph{Derived Features:} These features could either be derived from a member's profile (e.g., implied skills), or generated by an external algorithm (e.g., embedding for a member (\S\ref{subsec:representation})).
\end{itemize}
The retrieval service is built on top of LinkedIn's Galene search platform~\cite{galene_engine}, which handles the selection of candidates matching the query, and the initial scoring/pruning of these candidates, in a distributed fashion.

\item \textbf{Scoring/Ranking Service:} This component is responsible for the second-pass ranking of candidates corresponding to each query, and returning the ranked list of candidates to the front-end system for displaying in the product. Given a query, this service fetches the matching candidates, along with their features, from the retrieval service, and in parallel, computes the vector embedding for the query. Then, it performs the second-pass scoring of the candidates (which includes generation of similarity features based on query and member embeddings (\S\ref{subsec:representation})) and returns the top ranked results. The second-pass scoring can be performed either by a deep learning based model (\S\ref{subsec:deepmodel}), or any other machine learned model (e.g., a GBDT model, as discussed in \S\ref{subsec:currentmodels}), periodically trained and updated as part of an offline workflow (not shown in the figure).

\end{enumerate}

%% file: exp.tex
\section{Experiments}\label{sec:exp}
We next present the results from our offline experiments for the proposed models, and then discuss the trade-offs and design decisions to pick a model for online experimentation. We finally present the results of our online A/B test of the chosen model on \emph{LinkedIn Recruiter} product, which is based on unsupervised embeddings.

\subsection{Offline Experiments} \label{subsec:offline_exp}
To evaluate the proposed methodologies, we utilized LinkedIn Recruiter usage data collected over a two month period within 2017. This dataset consists of the impressions (recommended candidates) with tracked features from the candidates and recruiters, as well as the labels for each impression (positive/1 for impressions which resulted in the recruiter sending an inMail and the inMail being accepted, negative/0 otherwise). Furthermore, we filter the impression set for both training and testing sets to come from a random bucket, i.e., a subset of the traffic where the top $100$ returned search results are randomly shuffled and shown to the recruiters. The random bucket helps reduce positional bias \cite{joachims_2005}. We split training data and test data by time, which forms a roughly $70\%-30\%$ split. The dataset covers tens of thousands of recruiters, and millions of candidates recommended.

To evaluate the performance of the ranking models, we use offline replay, which re-ranks the recommended candidates based on the new model being tested, and evaluates the new ranked list. As explained previously, the main metric we report is precision at $k$ (Prec@$k$) due to its stability and the suitability with the way LinkedIn Recruiter product presents candidates. Prec@$k$ lift represents the \% gain in the inMail Accept precision for top $k$ impressions. 
Prec@$k$ is computed as the fraction of positive responses  (inMail accepts, within three days of the inMail being sent by the recruiter) in a given search session, averaged over all the sessions in the dataset. For training of the deep models, we utilized TensorFlow \cite{Abadi16}, an open-source software package for creating and training custom neural network models.

\input{exp_subsubsec_1}

\input{exp_subsubsec_2}

\input{exp_subsec_2}

%% file: exp_subsubsec_1.tex
\subsubsection{Deep Models} \label{subsubsec:exp_deep}
We first evaluated the effect of utilizing the end-to-end deep model, proposed in \S\ref{subsec:deepmodel}, with up to three layers on the dataset described above. The baseline model is a gradient boosted decision tree (\S\ref{subsec:currentmodels}, trained using the XGBoost \cite{xgboost} package) which consists of $30$ trees, each having a maximum depth of $4$ and trained in a point-wise manner, and we compare it to the neural network approach. The model family is a $k$ layer multi-layer perceptron (MLP) with $100$ units in each layer and rectified linear units (ReLU) for activations. We did not regularize the network because the size of the network was small enough, but rather used early stopping to achieve a similar effect. Also, as explained in \S\ref{subsec:deepmodel}, we chose hinge loss to train the pairwise training (the final metrics produced from logistic or hinge loss did not differ significantly).

\begin{table}[!htp]
\small
\caption{Precision lift of end-to-end MLP models trained with point-wise and pair-wise losses  as well as varying number of layers over the baseline gradient boosting tree model.}
\vspace{-0.17in}
\begin{tabular}{c|c|c|c|c}
	\hline\hline
	\textbf{Model} & \textbf{Optimization} & \textbf{Prec@1} & \textbf{Prec@5} & \textbf{Prec@25} \\\hline\hline
		XGBoost & - &0\% & 0\% & 0\% \\\hline
		1-layer & Pointwise & -2.93\% & -4.39\% & -1.72\% \\\hline
		3-layer & Pointwise & -0.31\% & -1.67\% & -0.19\% \\\hline
		1-layer & Pairwise  & -0.16\% & -1.36\%  & 0\% \\\hline
		3-layer & Pairwise & +5.32\% & +2.82\%  & +1.72\% \\\hline
	\end{tabular}
	\label{tab:deepOffline}
	\vspace{-0.12in}
\end{table}

The results are shown in Table~\ref{tab:deepOffline}. Interestingly, while single layer neural network trained with pointwise loss has poor ranking performance, additional layers of nonlinearity bring the neural network performance almost on par with XGBoost (further layers and units per layer did not improve the results, and are omitted here for brevity). On the other hand, neural network models trained using pairwise loss outperformed those trained with pointwise loss and XGBoost baseline as more layers are introduced (similar to the pointwise case, we did not see additional gains using more layers or units for pairwise loss). A possible explanation is the following:
\begin{enumerate}
\item Pairwise loss approach explicitly compares positive examples to negative examples within a search session rather than simply separate positive examples from negative examples in a broad sense, and,
\item It automatically deals with imbalanced classes since it could be mathematically shown that pairwise ranking loss is closely related to AUC \cite{gao2015consistency}, which is immune to class imbalance.
\end{enumerate}

%% file: exp_subsubsec_2.tex
\subsubsection{Shallow Models} \label{subsubsec:exp_embedding}
In this family of models, we use representation learning methods to construct dense vectors to represent certain categorical features. Although deep networks are used to train the embeddings, once trained, they are used as features in the baseline gradient boosted decision tree model, which is shallow.

Our first set of experiments utilizes unsupervised network embeddings, as proposed in \S\ref{subsec:representation} to learn representations for categorical variables like  skills and titles from member profiles. The title/skill representations are learned for both the query and the document (member) and a measure of similarity between the two is used as a new feature in the ranking model (baseline GBDT with additional feature). As shown in Table~\ref{tab:shallowOfflineUnsupervised}, converting the categorical interaction feature to a dense similarity measure results in large gains in the precision metric. The employed embedding was a concatenated version of order $1$ and order $2$ representations. For each member or query, the aggregation strategy used was mean pooling (although max pooling resulted in similar results), i.e., if a member or query has multiple skills, we do a mean pool of all the individual skill vectors to represent them on the vector space. Denote the mean pooled member vector by $m$, and the mean pooled query vector by $q$. We experimented with three similarity measures \cut{($s$) }between the two vector representations: 
\begin{enumerate}
\item \textbf{Dot Product:} $ m \bullet q = \sum_i m_i \cdot q_i$,
\item \textbf{Cosine Similarity:} $\frac{m \bullet q}{||m||_2 ~ ||q||_2} = \frac{\sum_i m_i \cdot q_i}{\sqrt{\sum_i m_i^2}\sqrt{\sum_i q_i^2}}$, and 
\item \textbf{Hadamard Product:} $m \circ q = \langle m_1 \cdot q_1, ~ \dots ~, m_d \cdot q_d \rangle$ (Also known as element-wise product).
\end{enumerate}
We note that both dot product and cosine similarity measures result in a single new feature added to the ranking model, whereas Hadamard product measure contributes to as many features as the dimensionality of the embedding vector representation. From Table~\ref{tab:shallowOfflineUnsupervised}, we can observe that using dot product outperformed using Hadamard product based set of features.

In our second set of experiments, we retain the same strategy of introducing feature(s) based on the similarity between the member/query embeddings into the ranking model. The only difference is that we now utilize a supervised algorithm (DSSM) to train the embeddings, which uses the same dataset as the offline experiments$~^3$\footnote{~$^3$ We first train the embeddings using DSSM on the training set explained in the beginning of \S\ref{subsec:offline_exp}. Then, we introduce the similarity measure based feature(s), and train the final ranking model based on GBDT on the same training dataset.}. As shown in Table \ref{tab:shallowOfflineSupervised}, we observed comparatively modest lift values in the Prec@k metric. In all experiments, we fixed the size of the embedding to $50$. We used tanh as the activation function and experimented with dot product and cosine similarity for the similarity computation between the two arms of DSSM. We used a minimum of 1 layer and a maximum of 3 layers in our experiments with DSSM models. The first hidden layer is used for word hashing, and the next two hidden layers are used to reduce the dimensionality of query / document vector representation. In our experiments, we did not observe better performance by using more than 3 layers. We conducted extensive offline experiments and tried over $75$ models. We only report the best configuration (network architecture) for each model.

\begin{table}[!htp]
\small
	\caption{Offline experiments with unsupervised embeddings.}
	\vspace{-0.19in}
	\begin{tabular}{c|c|c|c|c}
		\hline\hline
		\textbf{Model} & \textbf{Similarity} & \textbf{Prec@1} & \textbf{Prec@5} & \textbf{Prec@25} \\\hline\hline
		XGBoost & - & 0\% & 0\% & 0\% \\\hline
		Skill, Title & Dot & 2.71\% & 1.72\% & 1.06\% \\\hline
		Skill, Title & Hadamard & 0\% & 0.73\% & 0.36\% \\\hline
		Title & Dot & 2.31\% & 1.99\%  & 0.53\% \\\hline
		Skill & Dot & 2.05\% & -0.18\%  & -0.35\% \\\hline
	\end{tabular}
	\label{tab:shallowOfflineUnsupervised}
	\vspace{-0.20in}
\end{table}

\begin{table}[!htp]
\small
	\caption{Offline experiments using supervised embeddings. The network architecture is represented in square brackets. Only the best performing architecture type is shown for each dimension of evaluation (Similarity measure, Text vs. Facet)}
	\vspace{-0.19in}
	\begin{tabular}{c|c|c|c|c}
		\hline\hline
		\textbf{Model} & \textbf{Similarity}  & \textbf{Prec@1} & \textbf{Prec@5} & \textbf{Prec@25} \\\hline\hline
		XGBoost  & - & 0\% & 0\% & 0\% \\\hline
		Text [200, 100] & Dot & 3.62\% & -0.13\% & 0.15\% \\\hline
		Text [200, 100] & Cosine & 0.44\% & 0.55\% & -0.10\% \\\hline
		Text [500, 500, 128] & Dot & 0.55\% & -0.01\% & 0.38\% \\\hline
		 Title [500] & Dot & 2.42\% & -0.13\%  & -0.02\% \\\hline
	\end{tabular}
	\label{tab:shallowOfflineSupervised}
	\vspace{-0.19in}
\end{table}

%% file: exp_subsec_2.tex
\subsection{Online Experiments}\label{sec:onlineexp}
Based on the offline results, we have currently put off the online deployment of the end-to-end deep learning models in LinkedIn Recruiter due to the following reasons:
\begin{itemize}
\item There is a significant engineering cost to implementing end-to-end deep learning solutions in search systems since the number of items (candidates) that need to be scored can be quite large. Further, the relatively large amount of computation needed to evaluate deep neural networks could cause the search latency to be prohibitively large, especially when there are many candidates to be scored, thereby not meeting the real-time requirements of search systems such as LinkedIn Recruiter.
\item The offline evaluation for the end-to-end deep models (\S\ref{subsubsec:exp_deep}) showed an improvement of 1.72\% in Prec@25 for the 3-layer case, which, although impressive per our experience, does not currently justify the engineering costs discussed above.
\end{itemize}
Instead, we performed online A/B tests incorporating the unsupervised network embeddings (\S\ref{subsubsec:unsupervised_emb}) as a feature in the gradient boosted decision tree model. As in the case of the offline experiments, in the online setting, we first concatenate both the first and the second order embeddings for the search query, and for each potential candidate to be shown, then take the cosine similarity between the two concatenated embeddings, and use that as a single additional feature (to the baseline) in the gradient boosted decision tree (for both offline training and online model evaluation). Although the offline gain as evaluated in \S\ref{subsubsec:exp_embedding} is smaller compared to that of an end-to-end deep learning model, the engineering cost and computational complexity is much less demanding, since the embeddings can be precomputed offline, and the online dot product computation is relatively inexpensive compared to a neural network evaluation. An additional benefit of testing the embedding features with a tree model instead of an end-to-end deep model is that we can measure the impact of the new feature(s) in an apple-to-apple comparison, i.e., under similar latency conditions as the baseline model which is a tree model as well, with the embedding based feature as the only difference. Finally, we decided against deploying the supervised embeddings (\S\ref{subsubsec:supervised_emb}) due to relatively weaker offline experimental results (\S\ref{subsubsec:exp_embedding}).

The A/B test as explained above was performed on LinkedIn Recruiter users during the last quarter of 2017, with the control group being served results from the baseline model (gradient boosted decision tree model trained using XGBoost \cite{xgboost}) and the treatment group, consisting of a random subset of tens of thousands of users, being served results from the model that includes the unsupervised network embedding based feature. We present the results of the experiment in Table~\ref{tab:onlineExperiment}. Although the p-values are high for the experiments (as a result of relatively small sample sizes), we note that an increase of 3\% in the overall precision is an impressive lift, considering that precision is a hard metric to move, based on our domain experience.

\vspace{-0.1in}
\begin{table}[!htp]
\small
	\caption{Online A/B testing results. Comparing XGBoost model with vs. without network embedding based semantic similarity feature. }
	\vspace{-0.17in}
	\begin{tabular}{c|c|c|c}
		\hline\hline
		\textbf{ } & \textbf{Prec@5} & \textbf{Prec@25} & \textbf{Overall precision} \\\hline\hline
		Improvement & 2\% & 1.8\%  & 3\% \\ \hline
		p-value & 0.2 & 0.25 & 0.11 \\ \hline
	\end{tabular}
	\label{tab:onlineExperiment}
	\vspace{-0.22in}
\end{table}

%% file: lessons.tex
\subsection{Lessons Learned in Practice} \label{sec:lessons}
We next present the challenges encountered and the lessons learned as part of our offline and online empirical investigations. As stated in \S\ref{sec:onlineexp}, we had to weigh the potential benefits vs. the engineering cost associated with implementing end-to-end deep learning models as part of LinkedIn Recruiter search system. Considering the potential latency increase of introducing deep learning models into ranking, we decided against deploying end-to-end deep learning models in our system. Our experience suggests that hybrid approaches that combine offline computed embeddings (including potentially deep learning based embeddings trained offline) with simpler online model choices could be adopted in other large-scale latency-sensitive search and recommender systems. Such hybrid approaches have the following key benefits: (1) the engineering cost and complexity associated with computing embeddings offline is much lower than that of an online deep learning based system, especially since the existing online infrastructure can be reused with minimal modifications; (2) the latency associated with computing dot product of two embedding vectors is much lower than that of evaluating a deep neural network with several layers.

%% file: related.tex
\section{Related Work} \label{sec:related}
Use of neural networks on top of curated features for ranking is an established idea which dates back at least to \cite{caruana1996using}, wherein simple 2-layer neural networks are used for ranking the risk of mortality in a medical application. In \cite{burges_2005}, the authors use neural networks together with the logistic pairwise ranking loss, and demonstrate that neural networks outperform linear models that use the same loss function. More recently, the authors of \cite{cheng2016wide} introduced a model that jointly trains a neural network and a linear classifier, where the neural network takes dense features, and the linear layer incorporates cross-product features and sparse features.

Research in deep learning algorithms for search ranking has gained momentum especially since the work on Deep Structured Semantic Models (DSSM) \cite{Huang13}. DSSM involves learning the semantic similarity between a pair of text strings, where a sparse representation called tri-letter grams is used. The C-DSSM model \cite{Shen14} extends DSSM by introducing convolution and max-pooling after word hashing layer to capture local contextual features of words. The Deep Crossing model \cite{Shan16} focuses on sponsored search (ranking ads corresponding to a query), where there is more contextual information about the ads. Finally, two other popular deep ranking models that have been used for search ranking are the ARC-I \cite{Hu14} (a combination of  C-DSSM and Deep Crossing) and Deep Relevance Matching Model \cite{Guo16} (which introduces similarity histogram and query gating concepts).

There is extensive work on generating unsupervised embeddings. The notion of word embeddings (\textit{word2vec}) was proposed in \cite{Mikolov13}, inspiring several subsequent \textit{*2vec} algorithms. Several techniques have been proposed for graph embeddings, including classical approaches such as multidimensional scaling (MDS) \cite{Cox00}, IsoMap \cite{Tenenbaum00}, LLE \cite{Roweis00}, and Laplacian Eigenmap \cite{Belkin02}, and recent approaches such as graph factorization \cite{Ahmed13} and DeepWalk \cite{Perozzi14}. To generate embedding representation for the entities using the LinkedIn Economic Graph, we adopt \emph{Large-Scale Information Network Embeddings} approach~\cite{Tang15}, by changing how we construct the graph. While the graph used in~\cite{Tang15} considers the members of the social network as vertices, we instead define a weighted graph over the entities (e.g., skill, title, company), and use the number of members sharing the same entity on their profile to induce an edge weight between the vertices. Thus, we were able to reduce the size of the problem by a few orders of magnitude, thereby allowing us to scale the learning to all entities in the Economic Graph.
Finally, a recent study presents a unified view of different network embedding methods like LINE and node2vec as essentially performing implicit matrix factorizations, and proposes NetMF, a general framework to explicitly factorize the closed-form matrices that network embeddings methods including LINE and word2vec aim to approximate \cite{NetMF}.

%% file: conclusion.tex
\section{Conclusions and Future Work}\label{sec:conclusion}
In this paper, we presented our experiences of applying deep learning models as well as representation learning approaches for talent search systems at LinkedIn. We provided an overview of LinkedIn Recruiter search architecture, described our methodology for learning representations of sparse entities and deep models in the talent search domain, and evaluated multiple approaches in both offline and online settings. We also discussed challenges and lessons learned in applying these approaches in a large-scale latency-sensitive search system such as ours. 
Our design choices for learning semantic representations of entities at scale, and the deployment considerations in terms of weighing the potential benefits vs. the engineering cost associated with implementing end-to-end deep learning models should be of broad interest to academicians and practitioners working on large-scale search and recommendation systems.